%% file: main.tex
\documentclass{article}
\usepackage[preprint,nonatbib]{nips_2018}

\usepackage{graphicx}
\usepackage{amsmath}
\usepackage{bm}
\usepackage{amsfonts}
\usepackage{mathtools}
\usepackage[utf8]{inputenc}
\usepackage[hidelinks]{hyperref}
\usepackage{subcaption}
\usepackage[capitalize,noabbrev]{cleveref}
\usepackage{algorithm}
\usepackage{algpseudocode}
\usepackage{booktabs}
\usepackage{notoccite}
\usepackage{fancyhdr}
\usepackage{diagbox}
\usepackage{tabularx}

\DeclarePairedDelimiter\norm{\lVert}{\rVert}

\creflabelformat{equation}{#2\textup{#1}#3}

\begin{document}

\title{Generating Photo-Realistic Training Data to Improve Face Recognition Accuracy}

\author{
Daniel Sáez Trigueros, Li Meng \\
School of Engineering and Technology\\
University of Hertfordshire\\
Hatfield AL10 9AB, UK
\And
Margaret Hartnett \\
GBG plc\\
London E14 9QD, UK
}

\maketitle

\begin{abstract}
In this paper we investigate the feasibility of using synthetic data to augment face datasets. In particular, we propose a novel generative adversarial network (GAN) that can disentangle identity-related attributes from non-identity-related attributes. This is done by training an embedding network that maps discrete identity labels to an identity latent space that follows a simple prior distribution, and training a GAN conditioned on samples from that distribution. Our proposed GAN allows us to augment face datasets by generating both synthetic images of subjects in the training set and synthetic images of new subjects not in the training set. By using recent advances in GAN training, we show that the synthetic images generated by our model are photo-realistic, and that training with augmented datasets can indeed increase the accuracy of face recognition models as compared with models trained with real images alone.
\end{abstract}

\section{Introduction}
\label{sec:introduction}

Image synthesis is a widely studied topic in computer vision. In particular, face image synthesis has gained a lot of attention because of its diverse practical applications. These include facial image editing \cite{larsen2015autoencoding,yan2016attribute2image,yeh2016semantic,perarnau2016invertible,brock2016neural,ding2017exprgan,zhang2017age,zhou2017photorealistic,antipov2017face,shen2017learning,lu2017conditional,choi2017stargan,yin2017semi,shu2017neural,lample2017fader,he2017arbitrary}, face de-identification \cite{meden2017face,meden2018k,brkic2017know,wu2018privacy}, face recognition (e.g. data augmentation \cite{masi2016we,banerjee2017srefi,osadchy2017genface,crispell2017dataset,kortylewski2018training,mokhayeri2018domain} and face frontalisation \cite{zhu2013deep,zhu2014multi,hassner2015effective,zhu2015high,yim2015rotating,tran2017representation,huang2017beyond}) and artistic applications (e.g. video games and advertisements).

In this work, we focus on the applicability of face image synthesis for data augmentation. It is widely known that training data is one of the most important factors that affect the accuracy of deep learning models. The datasets used for training need to be large and contain sufficient variation to allow the resulting models to learn features that generalise well to unseen samples. In the case of face recognition, the datasets must contain many different subjects, as well as many different images per subject. The first requirement enables a model to learn inter-class discriminative features that can generalise to subjects not in the training set. The second requirement enables a model to learn features that are robust to intra-class variations. Even though there are several public large-scale datasets \cite{sun2014deep,yi2014learning,parkhi2015deep,guo2016ms,nech2017level,bansal2017umdfaces,cao2017vggface2} that can be used to train CNN-based face recognition models, these datasets are nowhere near the size or quality of commercial datasets. For example, the largest publicly available dataset contains about 10M images of 100K different subjects \cite{guo2016ms}, whereas Google's FaceNet \cite{schroff2015facenet} was trained with a private dataset containing between 100M and 200M face images of about 8M different subjects. Another issue is the presence of long-tail distributions in some publicly available datasets, i.e. datasets in which there are many subjects with very few images. Such unbalanced datasets can make the training process difficult and result in models that achieve lower accuracy than those trained with smaller but balanced datasets \cite{zhou2015naive}. In addition, some publicly available datasets (e.g. \cite{guo2016ms}) contain many mislabelled samples that can decrease face recognition accuracy if not discarded from the training set. Since collecting large-scale, good quality face datasets is a very expensive and labour-intensive task, we propose a method for generating photo-realistic face images that can be used to effectively increase the depth (number of images per subject) and width (number of subjects) of existing face datasets.

An approach that has recently gained popularity for augmenting face datasets is the use of 3D morphable models \cite{blanz2003face}. In this approach, new faces of existing subjects can be synthetized by fitting a 3D morphable model to existing images and modifying a variety of parameters to generate new poses and expressions \cite{masi2016we,crispell2017dataset,mokhayeri2018domain}. It is also possible to generate images with other variations using this approach. For example, \cite{mokhayeri2018domain} incorporated a reflectance model to generate images under different lighting conditions; and \cite{kortylewski2018training} randomly sampled 3D face shapes and colours to generate faces of new subjects. The main drawback of methods based on 3D morphable models is that the generated images often look unnatural and lack the level of detail found in real images. Another recent approach based on blending small triangular regions from different training images was proposed in \cite{banerjee2017srefi}. Although this method seemed to produce photo-realistic faces, the authors limited their work to frontal face images. In contrast, our approach makes use of generative adversarial networks (GANs) \cite{goodfellow2014generative}, which have recently been shown to produce photo-realistic in-the-wild images often indistinguishable from real images \cite{karras2017progressive}. Another advantage of using GANs is that they are end-to-end trainable models that do not require any domain{-specific processing, as opposed to methods based on 3D modelling or face triangulation.

Many methods based on GANs have been proposed for manipulating attributes of existing face images, including age \cite{antipov2017face,zhang2017age}, facial expressions \cite{zhou2017photorealistic,ding2017exprgan,choi2017stargan}, and other attributes such as hairstyle, glasses, makeup, facial hair, skin colour or gender \cite{larsen2015autoencoding,perarnau2016invertible,brock2016neural,shen2017learning,lu2017conditional,choi2017stargan,yin2017semi,shu2017neural,lample2017fader,he2017arbitrary}.
While these methods can be used to increase the depth of a dataset, it remains unclear how to increase the width of a dataset, i.e. how to generate faces of new subjects. Our proposed GAN is able to generate faces from a latent representation $\bm{z}$ that has two gaussian distributed components $\bm{z}_{id}$ and $\bm{z}_{nid}$ encoding identity{-related attributes and non-identity-related attributes respectively. In this way, face images of new subjects can be generated by fixing the identity component $\bm{z}_{id}$ and varying the non-identity component $\bm{z}_{nid}$. The method most closely related to ours is the \textit{semantically decomposed GAN} (SD-GAN) proposed in \cite{donahue2017semantically}. However, in contrast with SD-GAN, our method also supports the generation of face images of subjects that exist in the training set. In other words, our method can increase both the width and the depth of a given face dataset. Furthermore, our proposed GAN is arguably simpler to implement than SD-GAN and easier to incorporate into other GAN architectures.

To demonstrate the efficacy of our method, we trained several CNN-based face recognition models with different combinations of real and synthetic data. In most cases, the models trained with a combination of real and synthetic data outperformed the models trained with real data alone.

The rest of this paper is organised as follows. \cref{sec:related_work} provides a review of GAN methods related to ours. \cref{sec:proposed_method} explains each part of our proposed GAN and the loss functions used for training. \cref{sec:experiments} discusses our experimental results, both in terms of the quality of the synthetic images generated by our proposed GAN and the accuracy achieved by datasets augmented with the synthetic images. Finally, our conclusions are presented in \cref{sec:conclusions}.

\section{Related Work}
\label{sec:related_work}

The literature on face image synthesis is very extensive. In this section, we focus on GAN methods related to ours. For a recent survey on different face image synthesis methods (including GANs) see \cite{lu2017recent}.

GANs generate data by sampling from a probability distribution $p_{model}$ that is trained to match a true data generating distribution $p_{data}$. This is done by mapping a vector of random latent variables $\bm{z} \sim p_{\bm{z}}$ to a sample $G(\bm{z})$ through a generator network $G$, where $p_{\bm{z}}$ is a prior distribution that can be easily sampled (e.g. Gaussian or uniform). The generator is trained to fool a discriminator network $D$ that tries to determine whether a sample is real or generated (i.e. synthetic). Thus, the generator and discriminator are trained with opposing optimisation objectives. While the discriminator is trained to maximise the probability of correctly classifying both real and generated samples, the generator is trained to minimise the probability that the generated samples are classified as such. Formally, the standard GAN optimisation objective can be expressed as follows:
\input{equations/eq_gan_objective}
As training progresses, the discriminator gets better at distinguishing real from generated samples and the generator gets better at producing realistic samples that can fool the discriminator. The training is considered completed when the generator and the discriminator reach an equilibrium, i.e. when the generator and the discriminator stop improving. In practice, since GANs rarely reach an equilibrium, it is common to simply stop the training process whenever there is no noticeable improvement in the visual quality of the generated samples.

The training of GANs is often unstable and can lead to the mode collapse problem (this happens when the generator maps different values of $\bm{z}$ to the same output sample \cite{goodfellow2016nips}). Although some works have proposed heuristics that reduce this effect \cite{radford2015unsupervised,salimans2016improved}, a full understanding of the training dynamics of GANs remains an open research question. Based on the idea that optimising the training objective in \cref{eq:gan_objective} can be interpreted as minimising the Jensen-Shannon divergence between the true data generating distribution $p_{data}$ and the model distribution $p_{model}$ \cite{goodfellow2014generative}, the authors of \cite{arjovsky2017wasserstein} proposed a novel training objective for GANs that minimises the Wasserstein distance instead of the Jensen-Shannon divergence. This GAN variation, which was named the Wasserstein GAN (WGAN), was shown to be more stable and to reduce the mode collapse problem \cite{arjovsky2017wasserstein}. An improved formulation of this approach \cite{gulrajani2017improved} is considered one of the current state-of-the-art techniques for training GANs.

Many works on GANs have adopted a family of architectures known as \textit{deep convolutional GANs} (DCGANs) \cite{radford2015unsupervised}. DCGANs follow a set of guidelines that were proposed for stable training and good image quality. More recently, \cite{karras2017progressive} proposed a new methodology for training GANs that consist of progressively growing both the spatial resolution of real and generated images and the number of layers of the discriminator and generator networks (PGGAN). In this manner, the training is very stable at the start since low-resolution images are easier to generate than high-resolution images due to their lower dimensionality and hence diversity. As training progresses, and the resolution of the images is increased, the generator gradually learns to generate images with finer detail. In contrast, standard GANs that are tasked with learning high-resolution images from the outset are typically more unstable. Using the PGGAN approach together with several proposed heuristics, the authors of \cite{karras2017progressive} were able to generate impressive photo-realistic 1024 $\times$ 1024 images with a 5.4$\times$ speedup factor with respect to the standard GAN training approach.

Conditional versions of GANs allow the generation of samples with specific attributes. The first conditional GAN was introduced in \cite{mirza2014conditional} and consisted of feeding a label $\bm{y}$ encoding some attribute(s) of the data to both the generator and the discriminator. Using this approach, the input to the generator can be constructed by concatenating the vector of latent variables $\bm{z}$ with the label $\bm{y}$. Likewise, the input to the discriminator can be constructed by combining the input image $\bm{x}$ and the label $\bm{y}$ (for example, by upscaling $\bm{y}$ and appending it as an additional channel to $\bm{x}$). Several applications based on this idea have been proposed, such as facial attribute editing \cite{perarnau2016invertible,lu2017conditional}, facial expression manipulation \cite{ding2017exprgan} and face ageing \cite{zhang2017age,antipov2017face}. An alternative type of conditional GAN called auxiliary classifier GAN (AC-GAN) was proposed in \cite{odena2017conditional}. Instead of feeding the label $\bm{y}$ to the discriminator, AC-GANs use an auxiliary classifier in the discriminator that is tasked with predicting the label $\bm{y}$ that has been fed to the generator. AC-GANs have also been successfully used for various applications, including facial attribute editing \cite{he2017arbitrary}, face frontalisation \cite{tran2017representation} and multi-domain image-to-image translation \cite{choi2017stargan}. Moreover, the use of an auxiliary classifier to predict $\bm{y}$ was shown in \cite{odena2016semi,salimans2016improved} to improve image quality even when $\bm{y}$ was not fed to the generator.

In the case of face image synthesis, conditional GANs can generate images of subjects existing in the training set by considering the identity of a subject as an attribute $\bm{y}$. However, this approach does not allow generation of samples of new subjects. As far as we are aware, the SD-GAN method proposed in \cite{donahue2017semantically} is the only work that has attempted to solve this task. SD-GANs split the vector of latent variables $\bm{z}$ into two components $\bm{z}_I$ and $\bm{z}_O$ encoding identity-related attributes and non-identity-related attributes respectively. SD-GANs are trained with pairs of real images from the same subject and pairs of images generated with the same identity-related attributes $\bm{z}_I$ but different non-identity-related attributes $\bm{z}_O$. The discriminator learns to reject pairs of images when either they do not look photo-realistic or when they do not appear to belong to the same subject. Once the training is completed, since the latent variables $\bm{z}_I$ are forced to encode identity-related attributes, images of new subjects can be generated by feeding the generator with random $\bm{z}_I$ samples. Although SD-GANs have a similar goal to our proposed GAN, they cannot generate images of subjects in the training set. Our proposed GAN can generate images of subjects in the training set and uses a simpler architecture that does not need to be trained with pairs of images.

An approach related to conditional GANs is the InfoGAN model proposed in \cite{chen2016infogan}. The goal of an InfoGAN is to disentangle data attributes in an unsupervised way. This is done by maximising the mutual information \cite{cover2012elements} between a subset $\bm{c}$ of the latent variables $\bm{z}$ and the generated image $G(\bm{z})$. InfoGANs can be implemented with an auxiliary network in the discriminator that is trained to predict $\bm{c}$. As shown in \cite{chen2016infogan}, this method ensures that the latent variables $\bm{c}$ encode meaningful data attributes that are not lost during the generation process. Our proposed GAN incorporates elements of both conditional GANs and InfoGANs to disentangle identity-related attributes from non-identity-related attributes.

\section{Proposed Method}
\label{sec:proposed_method}

In this section, we first explain our choice of GAN architecture and type of conditional GAN, and then our proposed modifications to disentangle identity-related attributes from non-identity-related attributes.

\subsection{Conditional PGGAN}
\label{sub:conditional_pggan}

We follow the same architecture and training method proposed in the PGGAN work \cite{karras2017progressive} (we use the open-source code released by the authors in \cite{karras2017git} and keep the default training settings unless otherwise stated) and add our modifications to it. The training starts by generating 4 $\times$ 4 images. The number of layers in the generator and discriminator is then gradually increased from 1 to 11 (each time the resolution is doubled, two convolutional layers are added) until 128 $\times$ 128 images are generated. We do not generate higher resolution images because the network that we use for face recognition in our experiments takes 100 $\times$ 100 images as input. Following \cite{karras2017progressive}, instead of using the standard GAN objective shown in \cref{eq:gan_objective}, we use the WGAN training objective \cite{gulrajani2017improved}:
\input{equations/eq_wgan_objective}
where $\mathcal{D}$ is the set of 1-Lipschitz functions (for a full derivation of \cref{eq:wgan_objective} see \cite{arjovsky2017wasserstein}). When posed as a minimisation problem and adding a gradient penalty that enforces the Lipschitz constraint (the gradient of 1-Lipschitz functions are bounded to 1), the WGAN loss \cite{gulrajani2017improved} becomes:
\input{equations/eq_wgan_img_loss}
where $p_{\tilde{\bm{x}}}$ is a distribution of samples interpolated from generated samples $G(\bm{z})$ and real samples $\bm{x}$, and $\lambda$ is a weight controlling the contribution of the gradient penalty to the loss. Following \cite{gulrajani2017improved}, we set the value of $\lambda$ to 10.

To make this model conditional, we use the AC-GAN method, i.e. the generator is conditioned on identity labels $\bm{y}$ and an auxiliary network $D_c$ in the discriminator $D$ is trained to predict $\bm{y}$. Since the identity labels $\bm{y}$ are categorical, $D_c$ is trained as a classifier with cross-entropy-loss:
\input{equations/eq_acgan_crossentropy_loss}
Note that the generator is also trained to minimise this loss function so that the identity labels $\bm{y}$ are not ignored during the generation process.

\subsection{Identity Latent Space}
\label{sub:identity_latent_space}

The model described in \cref{sub:conditional_pggan} can generate images of subjects with identities $\bm{y}$ existing in the training set. However, it is not possible to generate images of subjects with new identities. For this reason, we propose the use of an embedding network $E$ to map the discrete identity labels $\bm{y}$ to a vector of latent variables $E(\bm{y})$. Since the goal is to learn a continuous latent space of identities that we can sample from, $E(\bm{y})$ is trained to follow a simple prior distribution $p_{\bm{z}_{id}}$ which in our case is a Gaussian. This can be done by using another discriminator $D_{\bm{z}_{id}}$ that is trained to match the posterior distribution defined by $E(\bm{y})$ to the Gaussian prior distribution $p_{\bm{z}_{id}}$ using adversarial training, as proposed in \cite{makhzani2015adversarial}. To incorporate the identity latent space into the conditional PGGAN model from \cref{sub:conditional_pggan}, the generator network $G$ is conditioned on the latent representation of the labels, i.e. $G(\bm{z} \mid E(\bm{y}))$. A diagram of our proposed GAN is shown in \cref{fig:proposed_gan}. The aforementioned modifications to the standard AC-GAN architecture are shown on the left side of \cref{fig:proposed_gan}.

We choose to train the embedding network $E$ to learn a stochastic mapping with Gaussian noise rather than a deterministic mapping. This is because in the deterministic case, $E$ can only use the stochasticity of the identity labels in the training set (which is fixed and typically very limited) to map the posterior distribution defined by $E(\bm{y})$ to the Gaussian prior distribution $p_{\bm{z}_{id}}$. Therefore, a deterministic mapping might not yield a smooth posterior distribution. In contrast, the additional randomness introduced by Gaussian noise in the stochastic mapping can alleviate this issue. With a stochastic mapping, the output of the embedding network $E$ is a vector of means and variances that is used to produce samples that must be indistinguishable from samples from the Gaussian prior distribution $p_{\bm{z}_{id}}$. In order to allow backpropagation through the sampling operation, the reparametrization trick proposed in \cite{kingma2013auto} is used.

To train the embedding network $E$ and the discriminator network $D_{\bm{z}_{id}}$, we again use the WGAN loss:
\input{equations/eq_wgan_embedding_loss}
where, in this case, $p_{\tilde{\bm{y}}}$ is a distribution of samples interpolated from the latent representation of the labels $E(\bm{y})$ and Gaussian samples $\bm{z}_{id}$. In a similar manner to \cref{eq:wgan_img_loss}, $\lambda_e$ is set to 10.

\subsection{Mutual Information Loss}
\label{sub:mutual_information_loss}

In our experiments, we observed that the increased dimensionality of $\bm{z}_{id}$ with respect to the discrete labels $\bm{y}$ might cause some or all of the non-identity-related attributes to be encoded by $\bm{z}_{id}$ instead of $\bm{z}_{nid}$. For this reason, we force $\bm{z}_{nid}$ to encode meaningful non-identity-related attributes by using a mutual information loss, as proposed in \cite{chen2016infogan}. As mentioned in \cref{sec:related_work}, this can be achieved through an auxiliary network $D_{mi}$ in the discriminator $D$ that is trained to predict $\bm{c}\subset\bm{z}_{nid}$. Since we do not have any prior knowledge about the latent variables $\bm{c}$, we treat them as continuous variables and train $D_{mi}$ as a regressor using minimum squared error (MSE):
\input{equations/eq_mi_mse_loss}

Balancing the mutual information loss from \cref{eq:mi_mse_loss} and the cross-entropy loss from \cref{eq:acgan_crossentropy_loss} is key to disentangling identity-related attributes from non-identity-related attributes.

\subsection{Proposed GAN}
\label{sub:proposed_gan}

\input{figures/fig_proposed_gan}

Referring to \cref{fig:proposed_gan}, it should be noted that, in practice, the auxiliary networks $D_c$ and $D_{mi}$ share all layers with the discriminator $D$. Hence, the last layer of $D$ is split into three components that are trained with different loss functions (adversarial loss $L_{img}$ for the real/generated classifier, cross-entropy loss $L_c$ for the identity classifier and MSE loss $L_{mi}$ for the non-identity-related attributes regressor). The architectures of the generator $G$ and the discriminator $D$ are the same as those in PGGAN \cite{karras2017progressive}. The embedding network $E$ contains an embedding layer that maps the discrete identity labels $\bm{y}$ to real-valued vectors, followed by two fully-connected layers. The discriminator network $D_{\bm{z}_{id}}$ contains three fully-connected layers. Both the dimensionality of the latent variables $\bm{z}_{id}$ encoding the identity-related attributes and the dimensionality of the latent variables $\bm{z}_{nid}$ encoding the non-identity-related attributes are fixed to 64. In our experiments, we did not notice any major difference between making $\bm{c}$ a subset of $\bm{z}_{nid}$ and simply making $\bm{c}$ equal to $\bm{z}_{nid}$. For simplicity, we adopted the latter option.

The proposed GAN is trained with the following overall loss:
\input{equations/eq_proposed_gan_loss}
where $\alpha$, $\beta$ and $\gamma$ are weights controlling the contributions of $L_c$, $L_e$ and $L_{mi}$ to the loss relative to the contribution of $L_{img}$. Note that \cref{eq:proposed_gan_loss} is the loss used when training the discriminators $D$ and $D_{\bm{z}_{id}}$. The generator $G$ and the embedding network $E$ are trained with the same loss as \cref{eq:proposed_gan_loss} except that the adversarial losses $L_{img}$ and $L_e$ have a negative sign. After extensive experimentation, we set $\alpha=1$, $\beta=1$ and $\gamma=50$. These weights are highly dependent on our specific architecture and should be tuned as necessary for different architectures.

Once the model is trained, we can generate multiple images of the same subject by feeding the generator with a fixed vector of identity-related attributes and different vectors of non-identity-related attributes. Our model allows generation of images of subjects in the training set by feeding the generator with the latent representation of their labels $E(\bm{y})$ obtained by mapping $\bm{y}$ through $E$, as shown in \cref{fig:proposed_gan_existing_id_generation}. Since $E(\bm{y})$ is trained to follow a Gaussian distribution $p_{\bm{z}_{id}}$, we can also feed the generator with a random sample $\bm{z}_{id}$ to generate images of new subjects, as shown in \cref{fig:proposed_gan_random_id_generation}.

\input{figures/fig_proposed_gan_existing_id_generation}
\input{figures/fig_proposed_gan_random_id_generation}

\section{Experiments}
\label{sec:experiments}

In this section, we start by providing a qualitative analysis of the synthetic images generated by our proposed GAN. Next, we explore the feasibility of augmenting face datasets with synthetic images, both in terms of width and depth. The augmented datasets are used to train CNN-based face recognition models (henceforth referred to as discriminative models) to determine whether they achieve a higher accuracy than models trained with real images alone.

Our discriminative models consist of a popular CNN architecture based on residual blocks that has been used in other face recognition works \cite{ranjan2017l2,hasnat2017deepvisage,wu2017deep}. The network is trained with softmax loss and optimised using stochastic gradient descent with momentum. The initial learning rate is set to 0.01 and decreased during training whenever the accuracy on the validation set stops improving. The input to the network are 100 $\times$ 100 RGB images. When training with datasets augmented with synthetic images we make sure that on each training batch the number of real and synthetic images is roughly the same.

\input{tables/tab_vggface_subsets}

We use three different subsets of the curated version of the VGGFace dataset \cite{parkhi2015deep} to train the discriminative models. The number of subjects and images of each subset is specified in \cref{tab:vggface_subsets}. We chose this dataset because it contains a good number of images per subject, which helps the training of our proposed GAN.

\subsection{Qualitative Analysis of Generated Images}
\label{sub:qualitative_analysis_of_generated_images}

We first train our proposed GAN using the VGGFace\textsuperscript{large} dataset. \cref{fig:generated_samples_existing_ids} shows synthetic images of subjects in the training set generated by our trained model using the method shown in \cref{fig:proposed_gan_existing_id_generation}. The identity-related attributes $E(\bm{y})$ have been fixed for each row and the non-identity-related attributes $\bm{z}_{nid}$ have been fixed for each column. The highlighted images in the first column are real images from the training set. Note how many of the synthetic images are as photo-realistic as the real images shown in the first column of \cref{fig:generated_samples_existing_ids}. \cref{fig:generated_samples_random_ids} shows synthetic images of new subjects generated by our trained model using the method shown in \cref{fig:proposed_gan_random_id_generation}. As in \cref{fig:generated_samples_existing_ids}, the identity-related attributes $\bm{z}_{id}$ have been fixed for each row and the non-identity-related attributes $\bm{z}_{nid}$ have been fixed for each column. Note how in both \cref{fig:generated_samples_existing_ids,fig:generated_samples_random_ids}, the images in each row appear to belong to the same subject since the identity-related attributes have been fixed. In contrast, the images in each column display common attributes that do not affect the identity of the subjects (e.g. head pose, facial expression and background) since the non-identity-related attributes have been fixed. From this we can conclude that our proposed GAN is able to effectively disentangle identity-related attributes from non-identity-related attributes.

\input{figures/fig_generated_samples_existing_ids}
\input{figures/fig_generated_samples_random_ids}

To test whether our method can generate images of subjects not present in the training set, we need to make sure that the synthetic images of new subjects indeed display identities that do not exist in the training set. \cref{fig:generated_samples_most_similar_a,fig:generated_samples_most_similar_b,fig:generated_samples_most_similar_c} show a comparison between synthetic images of three new subjects (shown in the top row of each of \cref{fig:generated_samples_most_similar_a,fig:generated_samples_most_similar_b,fig:generated_samples_most_similar_c}) and synthetic images of their most similar subject in the training set (shown in the bottom row of each of \cref{fig:generated_samples_most_similar_a,fig:generated_samples_most_similar_b,fig:generated_samples_most_similar_c}). These figures were created by measuring the average image difference between synthetic images of each new subject in \cref{fig:generated_samples_most_similar_a,fig:generated_samples_most_similar_b,fig:generated_samples_most_similar_c} and synthetic images of each subject in the training set. Since we were only interested in comparing the identity of the subjects, we averaged over image differences between synthetic images generated with the same non-identity-related attributes $\bm{z}_{nid}$, as observed in the columns of each pair of rows in \cref{fig:generated_samples_most_similar_a,fig:generated_samples_most_similar_b,fig:generated_samples_most_similar_c}. We can see how even though the synthetic images of new subjects look similar to the synthetic images of their most similar subject in the training set, it is possible to visually differentiate them as two different identities. Hence, we can conclude that our proposed GAN is able to successfully generate images of subjects not present in the training set.

\input{figures/fig_generated_samples_most_similar}

We can also show how our model has not overfit the training images by applying linear interpolation between two random vectors of identity-related attributes $\bm{z}_{id}$ and two random vectors of non-identity-related attributes $\bm{z}_{nid}$. \cref{fig:generated_samples_interpolation} shows how the transition between synthetic images generated from interpolated vectors is smooth and visually consistent with what would be expected from mixing attributes of two face images. This suggests that our model is able to generate images with enough diversity and it is not just learning to replicate the training images.

\input{figures/fig_generated_samples_interpolation}

\subsection{Augmenting Datasets with Synthetic Images}
\label{sub:augmenting_datasets_with_synthetic_images}

In order to evaluate the quality of datasets augmented with synthetic images, we train several discriminative models with different combinations of real and synthetic images and evaluate them against models trained with real images alone. Each augmented dataset is created by adding synthetic images to one of the VGGFace subsets shown in \cref{tab:vggface_subsets}. The synthetic images are generated using our proposed GAN trained with the same dataset that we want to augment. For example, if we want to augment VGGFace\textsuperscript{small}, we add synthetic images generated by our proposed GAN trained with VGGFace\textsuperscript{small}. In this way, we can realistically assess whether we can improve the performance of a discriminative model by augmenting its training set using our proposed method. All our discriminative models are evaluated using the IJB-A dataset \cite{klare2015pushing}. In particular, we use the verification protocol described in \cite{klare2015pushing} and report the true acceptance rate when the false acceptance rate is fixed to 0.01. We chose the IJB-A dataset for evaluation because it contains challenging images that do not overlap with any of the subjects in the VGGFace dataset.

\input{tables/tab_depth_augmented_accuracy}
\input{tables/tab_width_augmented_accuracy}

In \cref{tab:depth_augmented_accuracy} we show the accuracy of models trained with depth-augmented datasets, i.e. datasets augmented by increasing the number of images per subject with synthetic images. For each subject in the training set we generate multiple synthetic images by fixing the vector of identity-related attributes $E(\bm{y})$ and randomly sampling different vectors of non-identity-related attributes $\bm{z}_{nid}$. We can see how, in general, the accuracy of the models trained with depth-augmented datasets increases with respect to the models trained without synthetic images. In particular, we obtained maximum accuracy improvements of $+1.44\%$, $+2.22\%$ and $+4.51\%$ when adding 500 synthetic images per subject to the VGGFace\textsuperscript{large}, VGGFace\textsuperscript{medium} and VGGFace\textsuperscript{small} datasets respectively. These results are consistent with the intuition that adding synthetic images to smaller datasets should result in greater improvement than adding them to larger datasets which contain more real images. The results shown in \cref{tab:depth_augmented_accuracy} also suggest that there is an optimal balance between the number of real and synthetic images per subject in a given dataset. Indeed, adding 1,000 synthetic images per subject to the VGGFace\textsuperscript{large}, VGGFace\textsuperscript{medium} and VGGFace\textsuperscript{small} datasets resulted in lower accuracy than adding 500 synthetic images per subject since the proportion of real images per subject becomes smaller.

In \cref{tab:width_augmented_accuracy} we show the accuracy of models trained with width-augmented datasets, i.e. datasets augmented by increasing the number of subjects with synthetic images of new subjects. For each new subject, we generate 500 synthetic images (since this was the best number of synthetic images per subject obtained in \cref{tab:depth_augmented_accuracy}) by fixing a randomly sampled vector of identity-related attributes $\bm{z}_{id}$ and randomly sampling different vectors of non-identity-related attributes $\bm{z}_{nid}$. Again, we observe improvement in most cases. In particular, we obtained a maximum accuracy improvement of $+1.19\%$ when adding 1,500 synthetic subjects to the VGGFace\textsuperscript{large} dataset; and $+4.23\%$ and $+8.42\%$ when adding 1,000 synthetic subjects to the VGGFace\textsuperscript{medium} and VGGFace\textsuperscript{small} datasets respectively. In this case, we also observe that adding synthetic images to the VGGFace\textsuperscript{large} dataset does not significantly change the recognition accuracy. This can be explained by the fact that this dataset already contains a large number of real subjects. In contrast, we observe a large improvement when increasing the number of synthetic subjects in the VGGFace\textsuperscript{medium} and VGGFace\textsuperscript{small} datasets, as the number of real subjects in neither of these datasets is very large. We also observe that there seems to be an optimal balance between the number of real and synthetic subjects in a given dataset. Indeed, as shown in \cref{tab:width_augmented_accuracy}, adding 1,500 synthetic subjects to the VGGFace\textsuperscript{medium} and VGGFace\textsuperscript{small} does not result in higher accuracy than adding 1,000 synthetic subjects since the proportion of real subjects becomes smaller. Note that in the case of the VGGFace\textsuperscript{large} dataset, more synthetic subjects can be added since there is still a good balance between real and synthetic subjects. However, as mentioned earlier, this dataset already contains a large number of real subjects. Hence, it is expected that no significant improvement in recognition accuracy will be obtained by adding more synthetic subjects.

Looking at the results shown in \cref{tab:depth_augmented_accuracy,tab:width_augmented_accuracy}, we can conclude that augmenting datasets with synthetic images is mainly beneficial for small and medium datasets. Moreover, the accuracy improvement obtained when training with width-augmented and depth-augmented datasets is relative to the number of subjects and number of images per subject of each augmented dataset. For example, the VGGFace\textsuperscript{small} dataset contains 200 subjects and an average of 295 images per subject. Thus, it is reasonable that the accuracy is improved by a larger margin when adding synthetic images of new subjects ($+8.42\%$) than when adding synthetic images of existing subjects ($+4.51\%$).

\section{Conclusions}
\label{sec:conclusions}

In this paper, we have studied the feasibility of augmenting face datasets with photo-realistic synthetic images. In particular, we have presented a new type of conditional GAN that can generate photo-realistic face images from two latent vectors encoding identity-related attributes and non-identity-related attributes respectively. By fixing the latent vector of identity-related attributes and varying the latent vector of non-identity-related attributes, our proposed GAN can generate images of subjects with fixed identities but different attributes, such as facial expression and head pose. The introduction of an embedding network to map discrete identity labels to a continuous latent space of identities allows us to both generate images of subjects in the training set and generate images of new subjects not in the training set. Our experiments have shown the effectiveness of the disentangled representation and the high visual quality of the generated images. To demonstrate the benefit of augmenting datasets with our method, we have trained several CNN-based face recognition models with different combinations of real and synthetic images. In most cases, the discriminative models trained with a combination of real and synthetic images have outperformed the discriminative models trained with real images alone. According to our experimental results, our method is particularly effective when augmenting datasets with a moderate number of subjects and/or images per subject.

Compared to other face image synthesis methods explicitly designed to generate face images, our method is more generic and can be used to generate any kind of images. Moreover, by only adding a few simple modifications to the standard AC-GAN architecture, our method can be easily extended. For example, in our particular case, the proposed GAN could be extended to control the non-identity-related attributes explicitly by conditioning the GAN on specific attributes. This would allow face datasets to be augmented in a tailored manner, e.g. by adding synthetic images of subjects with sunglasses, facial hair, different ages, etc. We hope that this and other ideas derived from our work will contribute to the development of new data augmentation techniques.

\bibliographystyle{ieeetr}
\bibliography{bibliography}

\end{document}

%% file: equations/eq_gan_objective.tex
\begin{equation}
    \min_{G}\max_{D}\mathbb{E}_{\bm{x} \sim p_{data}} [\log D(\bm{x})] + \mathbb{E}_{\bm{z} \sim p_{\bm{z}}} [\log (1-D(G(\bm{z})))]
    \label{eq:gan_objective}
\end{equation}

%% file: equations/eq_wgan_objective.tex
\begin{equation}
    \min_{G}\max_{D\in\mathcal{D}}\mathbb{E}_{\bm{x} \sim p_{data}} [D(\bm{x})] - \mathbb{E}_{\bm{z} \sim p_{\bm{z}}} [D(G(\bm{z}))]
    \label{eq:wgan_objective}
\end{equation}

%% file: equations/eq_wgan_img_loss.tex
\begin{equation}
    L_{img} = \mathbb{E}_{\bm{z} \sim p_{\bm{z}}} [D(G(\bm{z}))] - \mathbb{E}_{\bm{x} \sim p_{data}} [D(\bm{x})] + \lambda ~ \mathbb{E}_{\tilde{\bm{x}} \sim p_{\tilde{\bm{x}}}} [\norm{(\nabla_{\tilde{\bm{x}}}D(\tilde{\bm{x}})}_2-1)^2]
    \label{eq:wgan_img_loss}
\end{equation}

%% file: equations/eq_acgan_crossentropy_loss.tex
\begin{equation}
    L_{c} = -\mathbb{E}_{\bm{z} \sim p_{\bm{z}},\bm{y} \sim p_{\bm{y}}} [\log D_c(G(\bm{z}\mid\bm{y}))] - \mathbb{E}_{\bm{x} \sim p_{data}} [\log D_c(\bm{x})]
    \label{eq:acgan_crossentropy_loss}
\end{equation}

%% file: equations/eq_wgan_embedding_loss.tex
\begin{equation}
    L_{e} = \mathbb{E}_{\bm{y} \sim p_y} [D_{\bm{z}_{id}}(E(\bm{y}))] - \mathbb{E}_{\bm{z}_{id} \sim p_{\bm{z}_{id}}} [D_{\bm{z}_{id}}(\bm{z}_{id})] + \lambda_e ~ \mathbb{E}_{\tilde{\bm{y}} \sim p_{\tilde{\bm{y}}}} [\norm{(\nabla_{\tilde{\bm{y}}}D_{\bm{z}_{id}}(\tilde{\bm{y}})}_2-1)^2]
    \label{eq:wgan_embedding_loss}
\end{equation}

%% file: equations/eq_mi_mse_loss.tex
\begin{equation}
    L_{mi} = \mathbb{E}_{\bm{z}_{nid} \sim p_{\bm{z}_{nid}},\bm{y} \sim p_{\bm{y}}} [\norm{\bm{c} - D_{mi}(G(\bm{z}_{nid} \mid E(\bm{y})))}_2^2]
    \label{eq:mi_mse_loss}
\end{equation}

%% file: figures/fig_proposed_gan.tex
\begin{figure}[tb]
    \centering
    \includegraphics[width=\linewidth]{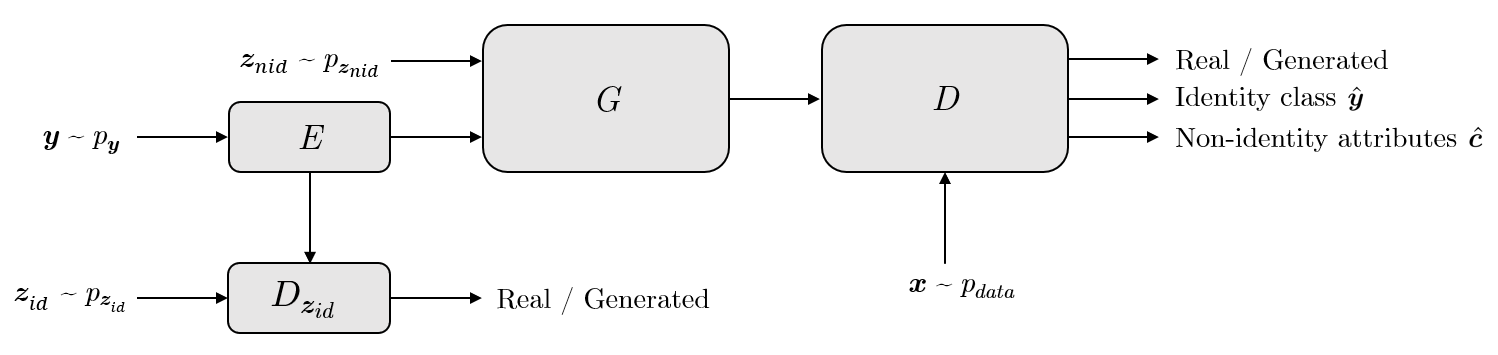}
    \caption{Proposed GAN model.}
    \label{fig:proposed_gan}
\end{figure}

%% file: equations/eq_proposed_gan_loss.tex
\begin{equation}
    L = L_{img} + \alpha L_c + \beta L_e + \gamma L_{mi}
    \label{eq:proposed_gan_loss}
\end{equation}

%% file: figures/fig_proposed_gan_existing_id_generation.tex
\begin{figure}[tb]
    \centering
    \includegraphics[width=0.55\linewidth]{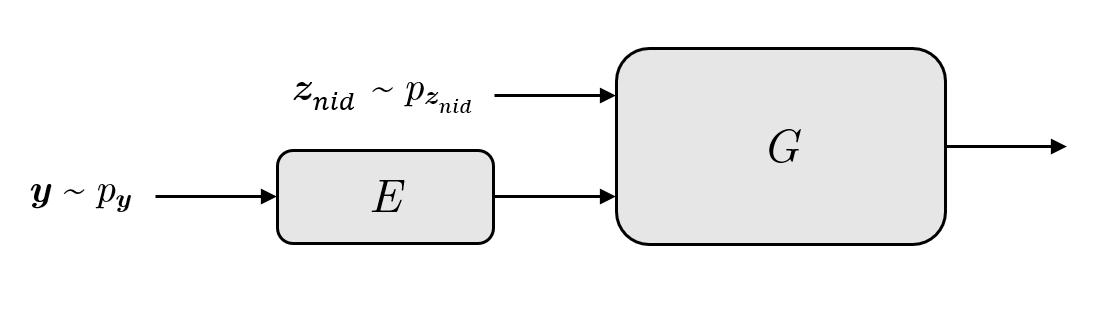}
    \caption{Generation of images of subjects $\bm{y}$ in the training set with identity-related attributes $E(\bm{y})$ and random non-identity-related attributes $\bm{z}_{nid}$.}
    \label{fig:proposed_gan_existing_id_generation}
\end{figure}

%% file: figures/fig_proposed_gan_random_id_generation.tex
\begin{figure}[tb]
    \centering
    \includegraphics[width=0.43\linewidth]{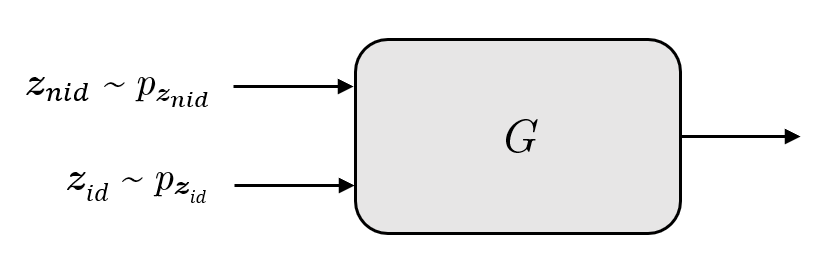}
    \caption{Generation of images of new subjects with random identity-related-attributes $\bm{z}_{id}$ and random non-identity-related attributes $\bm{z}_{nid}$.}
    \label{fig:proposed_gan_random_id_generation}
\end{figure}

%% file: tables/tab_vggface_subsets.tex
\begin{table}[b]
    \centering
    \caption{VGGFace subsets used for training our models.}
    \label{tab:vggface_subsets}
    \begin{tabular}{@{}lcc@{}}
    \toprule
    Dataset                          & Number of subjects & Number of images \\ \midrule
    VGGFace\textsuperscript{large}   & 2558               & 734,665          \\
    VGGFace\textsuperscript{medium}  & 800                & 227,466          \\
    VGGFace\textsuperscript{small}   & 200                & 58,952           \\ \bottomrule
    \end{tabular}
\end{table}

%% file: figures/fig_generated_samples_existing_ids.tex
\begin{figure}[tb]
    \centering
    \includegraphics[width=\linewidth]{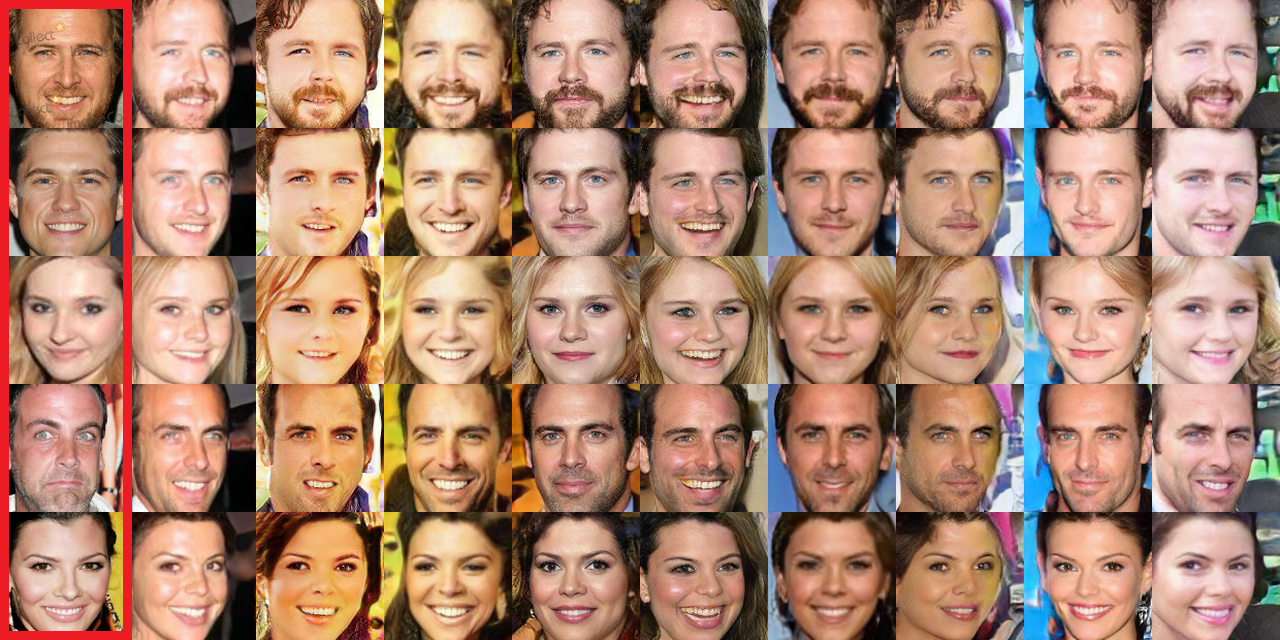}
    \caption{Synthetic images of subjects in the training set generated by our proposed GAN using the method shown in \cref{fig:proposed_gan_existing_id_generation}. The identity related attributes have been fixed for each row and the non-identity related attributes have been fixed for each column. Note that the highlighted images in the first column are real images from the training set.}
    \label{fig:generated_samples_existing_ids}
\end{figure}

%% file: figures/fig_generated_samples_random_ids.tex
\begin{figure}[tb]
    \centering
    \includegraphics[width=\linewidth]{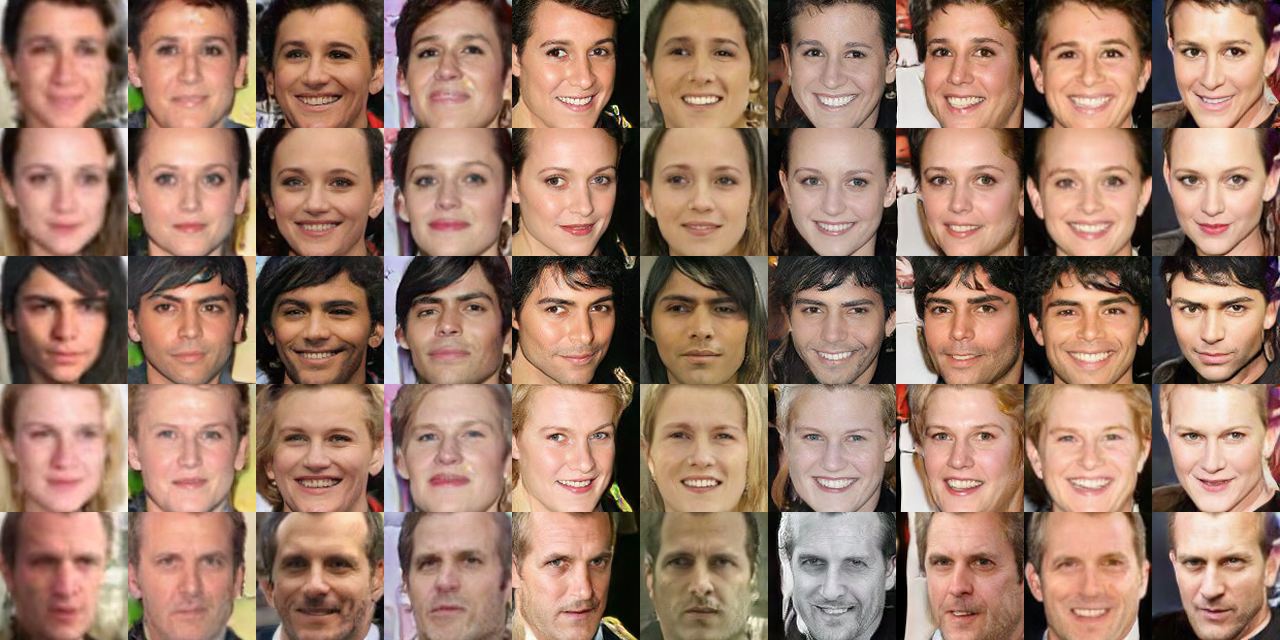}
    \caption{Synthetic images of new subjects generated by our proposed GAN using the method shown in \cref{fig:proposed_gan_random_id_generation}. The identity related attributes have been fixed for each row and the non-identity related attributes have been fixed for each column.}
    \label{fig:generated_samples_random_ids}
\end{figure}

%% file: figures/fig_generated_samples_most_similar.tex
\begin{figure}[tb]
    \centering
    \subcaptionbox{\label{fig:generated_samples_most_similar_a}}{\includegraphics[width=\linewidth]{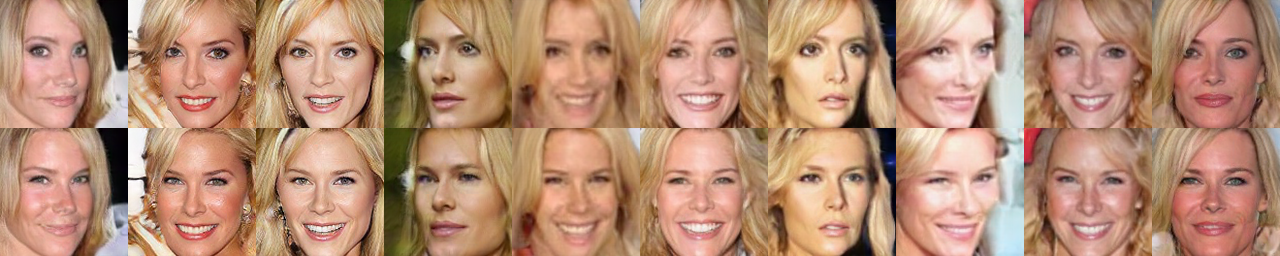}}
    \par\medskip
    \subcaptionbox{\label{fig:generated_samples_most_similar_b}}{\includegraphics[width=\linewidth]{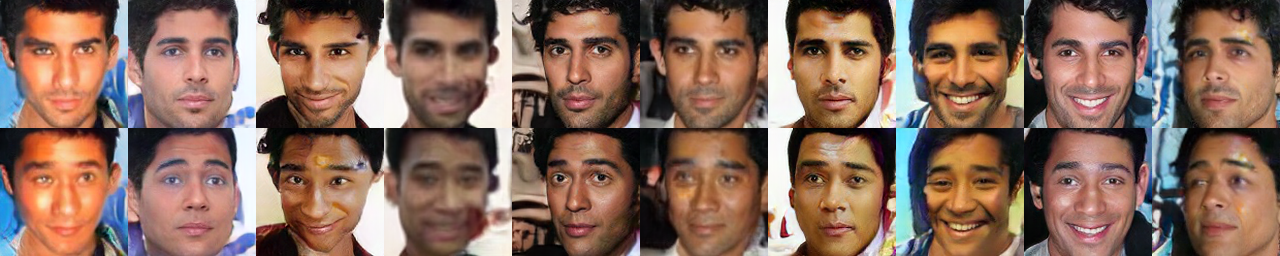}}
    \par\medskip
    \subcaptionbox{\label{fig:generated_samples_most_similar_c}}{\includegraphics[width=\linewidth]{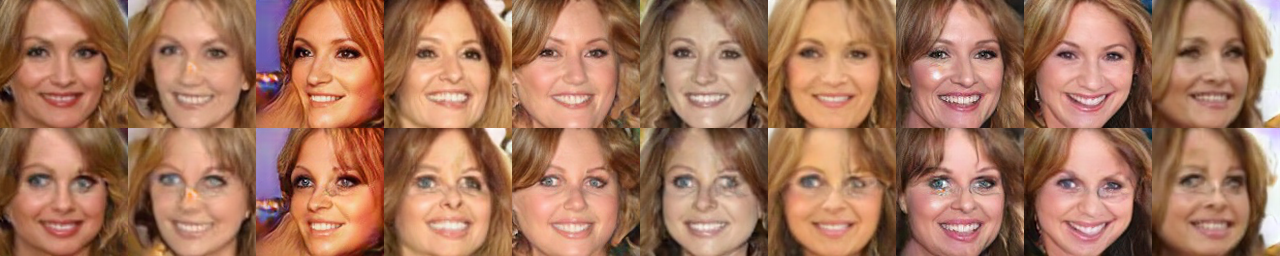}}
    \caption{Comparison between synthetic images of new subjects and synthetic images of their most similar subject in the training set. The top row of each of \subref*{fig:generated_samples_most_similar_a}, \subref*{fig:generated_samples_most_similar_b}, \subref*{fig:generated_samples_most_similar_c} contains synthetic images of a new subject and the bottom row of each of \subref*{fig:generated_samples_most_similar_a}, \subref*{fig:generated_samples_most_similar_b}, \subref*{fig:generated_samples_most_similar_c} contains synthetic images of their most similar subject in the training set. Note that the non-identity-related attributes only vary across the rows of \subref*{fig:generated_samples_most_similar_a}, \subref*{fig:generated_samples_most_similar_b}, \subref*{fig:generated_samples_most_similar_c} to restrict the comparison to the identity of the subjects.}
    \label{fig:generated_samples_most_similar}
\end{figure}

%% file: figures/fig_generated_samples_interpolation.tex
\begin{figure}[tb]
    \centering
    \includegraphics[width=0.56\linewidth]{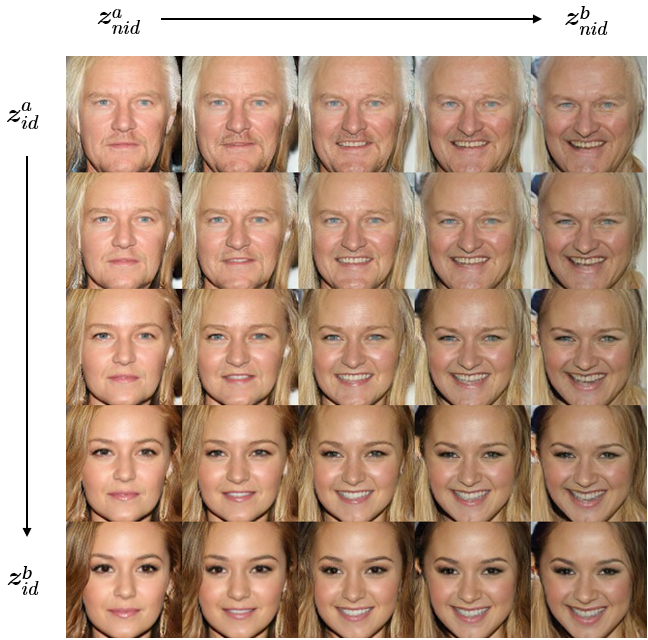}
    \caption{Synthetic images generated by interpolating between two random vectors of identity related attributes $\bm{z}_{id}^a$, $\bm{z}_{id}^b$ and two random vectors of non-identity related attributes $\bm{z}_{nid}^a$, $\bm{z}_{nid}^b$.}
    \label{fig:generated_samples_interpolation}
\end{figure}

%% file: tables/tab_depth_augmented_accuracy.tex
\begin{table}[b]
    \centering
    \caption{Accuracy of discriminative models trained with depth-augmented datasets. The reported accuracy corresponds to the TAR@FAR=0.01 obtained when evaluating the models on the IJB-A dataset.}
    \label{tab:depth_augmented_accuracy}
    \begin{tabularx}{0.65\textwidth}{l*{4}{>{\centering\arraybackslash}X}}
    \toprule
    & \multicolumn{4}{c}{Number of synthetic images per subject} \\
    \cmidrule(l){2-5}
    Training set                     & 0       & 250     & 500     & 1000    \\ \midrule
    VGGFace\textsuperscript{large}   & 67.58\% & 66.65\% & $\boldsymbol{69.02\%}$ & 67.74\% \\
    VGGFace\textsuperscript{medium}  & 50.32\% & 52.25\% & $\boldsymbol{52.54\%}$ & 51.97\%         \\
    VGGFace\textsuperscript{small}   & 30.64\% & 33.30\% & $\boldsymbol{35.15\%}$ & 32.95\% \\ \bottomrule
    \end{tabularx}
\end{table}

%% file: tables/tab_width_augmented_accuracy.tex
\begin{table}[tb]
    \centering
    \caption{Accuracy of discriminative models trained with width-augmented datasets. The reported accuracy corresponds to the TAR@FAR=0.01 obtained when evaluating the models on the IJB-A dataset.}
    \label{tab:width_augmented_accuracy}
    \begin{tabularx}{0.65\textwidth}{l*{4}{>{\centering\arraybackslash}X}}
    \toprule
    & \multicolumn{4}{c}{Number of synthetic subjects} \\
    \cmidrule(l){2-5}
    Training set                     & 0       & 500     & 1000     & 1500    \\ \midrule
    VGGFace\textsuperscript{large}   & 67.58\% & 65.76\% & 66.32\% & $\boldsymbol{68.77\%}$ \\
    VGGFace\textsuperscript{medium}  & 50.32\% & 52.15\% & $\boldsymbol{54.55\%}$ & 54.32\% \\
    VGGFace\textsuperscript{small}   & 30.64\% & 38.06\% & $\boldsymbol{39.06\%}$ & 38.81\% \\ \bottomrule
    \end{tabularx}
\end{table}